# SlicerOrbitSurgerySim: An Open-Source Platform for Virtual Registration and Quantitative Comparison of Preformed Orbital Plates

**Chi Zhang,**

Department of Biomedical Sciences,

Texas A&M University College of Dentistry, Dallas, TX, USA

**Braedon Gunn,**

Texas A&M University College of Dentistry, Dallas, TX, USA

**Andrew M. Read-Fuller,**

Department of Oral and Maxillofacial Surgery,

Texas A&M University College of Dentistry, Dallas, TX, USA

**Abstract**

Poor adaptation of orbital implants remains a major contributor to postoperative complications and revision surgery. Although preformed orbital plates are widely used to reduce cost and operative time compared with customized implants, surgeons currently lack publicly available tools and standardized metrics to quantitatively compare plate fit across vendors, sizes, and patient anatomy. We developed SlicerOrbitSurgerySim, an open-source extension for the 3D Slicer platform that enables interactive virtual registration, evaluation, and comparison of multiple preformed orbital plates in a patient-specific virtual planning environment. The software generates reproducible quantitative plate-to-orbit distance metrics and visualization tools that support both patient-specific planning and population-level statistical analysis of plate adaptability. By facilitating objective comparison of implant designs and placement strategies, this tool aims to improve preoperative decision-making, reduce intraoperative plate modification, and promote collaborative research and surgical education. Pilot studies, sample datasets, and detailed tutorials are provided to support testing, transparency, and reproducibility.

**Innovation**

In orbital fracture surgery, preformed plates are frequently used to reduce the cost and time of customization[1,2]. The only available software for craniofacial trauma planning is proprietary, focusing on generating customized plans for plate placement and modification (e.g., bending or cutting) for intraoperative navigation[1,3–7]. These workflows are *ad hoc* and rely heavily on vendor-employed engineers[3,4]. As a result, there is no standardized, reproducible, and accessible method for surgeons to quantitatively compare the fit between plates of different sizes and from multiple vendors and patient-specific orbital anatomy[3,4,8]. This makes it difficult to select the ideal plate design or placement strategies in advance, which may increase intraoperative plate modification time and impact fit and postoperative clinical outcomes.

To address this gap, we developed *SlicerOrbitSurgerySim*, an open-source software extension of the 3D Slicer medical imaging platform[9] that allows surgeons to virtually register multiple preformed plates, compute plate-to-orbit distance metrics, and visualize global and regional fit. To our knowledge, this is the first toolkit designed specifically for multi-plate, multi-vendor, surgeon-directed comparison of preoperative plate fit analysis. We also present an end-to-end workflow, spanning image segmentation, orbital reconstruction, plate registration, and quantitative comparison, within a single platform.

We demonstrate the utility of this tool through a pilot study of four plate types from two vendors. The resulting quantitative metrics support both downstream statistical analyses and case-specific comparisons, providing insight into how orbital morphology interacts with plate design and assisting in the development of plate-placement strategies.

**Advantages**

Conventional proprietary systems are designed to plan for intraoperative placement of a single customized plate[1,2,6,7,10]. These systems do not support comparing multiple preformed plates from different vendors preoperatively or allow the surgeon to control the software directly. One prior study adapted commercial software to visualize several plates in superimposition, but the assessment remained qualitative and limited to postoperative evaluation[3]. In addition, the high cost and closed nature of proprietary systems restrict collaboration, innovation, and research[3,4,8].

*SlicerOrbitSurgerySim* addresses these limitations by generating standardized, quantitative fit metrics, i.e., deviations between the plate and reconstructed orbit, entirely preoperatively and virtually. Surgeons can compare multiple plates across vendors, assess different placement strategies, and evaluate both overall and regional deviations between plates and orbits. Versatile visualization tools allow detailed assessment of global and localized adaptation.

The software allows interactive, iterative adjustment of plate positions and customization of fit metrics. Users can simulate alternative placement strategies, including trimming and repositioning, and immediately evaluate their impact on plate fit.

Because the module is built within the 3D Slicer ecosystem, it integrates seamlessly with many 3D imaging and modeling tools, including AI-based automated segmentation. We also developed a complementary module for orbital reconstruction using the mirrored contralateral orbit, which provides a consistent reference for evaluating how closely a plate conforms to the reconstructed orbital shape.

Although the software requires basic familiarity with CT segmentation and the 3D Slicer interface, step-by-step tutorials and sample datasets reduce the learning curve. Together, these features support both clinical use and hypothesis-driven research in virtual orbital reconstruction.

**Significance**

Orbital surgery requires meticulous planning as minor changes in orbital contents can cause discomfort, functional deficit, and aesthetic issues, such as enophthalmos, limited ocular motion, and diplopia[1,2]. Orbital reconstruction typically involves ordering a customized implant or utilizing a preformed orbital plate with or without modification (e.g., bending and trimming) [1–3,6,11,12]. Despite advances in plate manufacturing and navigation technology, plate misfit, including malpositioning, and deformation, continues to contribute to postoperative complications in orbital surgery [1,2,4–6,8]. One study found surgical revision cases (12 of 71 or 17% of the study cohort) involved both plate malposition and associated soft tissue complications [4]. However, "malposition" or "inaccurate position" in previous studies have been loosely defined, often based on qualitative visual comparison of postoperative CT contours or on deviation from a planned plate position rather than on objective metrics [1–4,8,11,13]. This ambiguity makes it difficult to determine whether a malpositioned implant reflects limitations in plate design, suboptimal placement, fracture-related constraints, or individual orbital morphology [2,3,8]. Though infrequent at the population level, each case represents a significant burden for the patient and clinicians.

Despite its importance, the field still lacks data-driven research on how plate designs affect surgical practice and outcome, such as how different plate sizes and contours influence modification and placement strategies or whether certain geometries specific patient groups to poorer fit[3,4]. Consequently, surgeons have limited guidance when selecting the most suitable plate among various types and vendors.

Several factors have impeded development of reference systems for plate selection. Proprietary software is costly and inaccessible, restricting adapting technologies for research purposes, developing standardized plate-fit metrics, and making cross-institutional or multi-operator research difficult[3,4]. Ethical considerations and the invasive nature of orbital surgery also limit opportunities to collect intraoperative or postoperative plate-fit data at scale[3]. As a result, no widely available framework exists for studying plate–orbit conformity systematically or for establishing quantitative approaches to guide plate choice and placement strategies.

*SlicerOrbitSurgerySim* allows low-cost, open, and repeatable evaluation of plate-orbit fit in a virtual environment. The tool generates reproducible plate registrations and standardized conformity metrics that can be shared across operators and institutions. These data can help surgeons and researchers examine how plate design interacts with orbital anatomy, explore covariates of fit such as gender or fracture pattern. Collectively, this supports the development of a quantitative framework for plate selection, improves preoperative decision-making, reduces intraoperative plate modification, and promotes collaborative research and surgical education. This establishes a foundation for future work on predicting surgical outcomes and refining implant designs.

By making this tool openly accessible, the software empowers surgeons to engage directly in virtual planning rather than relying solely on vendor-employed engineers. In the future, this model may further lower technological barriers to using 3D imaging and virtual planning tools, accelerating technological innovation not only in orbital reconstruction, but in other areas of maxillofacial surgery as well.

**Evidence**

The *SlicerOrbitSurgerySim* toolkit can be installed directly through the 3D Slicer Extension Manager. Detailed step-by-step tutorials are provided in the software repository. After installation, sample data can be accessed through Slicer's "Sample Data" module.

**A. Workflow summary**

The entire workflow was performed in 3D Slicer (**Figure 1**; see the tutorial for detail).

The CT images were segmented to produce 3D models of the bony orbit. To approximate each patient's pre-injury anatomy, the fractured orbit was reconstructed by mirroring the contralateral side, a widely used approach in orbital reconstruction planning[1,2,6–8]. The *MirroredOrbitRecon* module (utilizing SlicerMorph[14,15] functions) from the *SlicerOrbitSurgerySim* performs rigid alignment between mirrored and fractured sides and offers optional CPD refinement to reduce natural left–right asymmetry (**Figure 2**).

Each plate was then registered to each orbit using the *PlateRegistration* module. The workflow consisted of:

1. Landmark-based initial positioning (**Figure 3A**)
2. Posterior-stop alignment (**Figure 3B**)

3. Manual fine adjustment using Slicer's transform handle with collision checking (**Figure 3C-3F**)

4. Generate fit metrics (**Figure 4**)

5. Fit comparison and ranking **(see the [tutorial](tutorial))**.

All plate registration results were verified by a board-certified oral and maxillofacial surgeon (AMR).

A key element of this workflow is anchoring the plate at the posterior stop, corresponding to the orbital process of the palatine bone and usually serves as the posterior fixation point[12] (**Figure 3B**). After coarse landmark alignment, the plate is refined by manually rotating around this posterior stop, reflecting how surgeons adjust the plate while ensuring posterior stop alignment (**Figure 3C**). Collision detection and visualization tools assists placing the plate just at the surface the unfractured peripheral orbital bone with no intersection for simulating realistic placement (**Figure 3D and 3E**).

**B. Fit metrics and ranking**

Plate fit was quantified as distances in mm between each registered plate and the reconstructed orbit. Smaller distances indicate closer conformity to the patient's pre-injury orbital contour using the reconstructed orbit as a proxy.

Two complementary metrics were used:

1. **Plate-wide distances** are pointwise distances from plate model vertices to the nearest point on the reconstructed orbit and can be visualized as heatmaps to assess global and regional conformity.

2. **Edge-specific regional distances** measures deviations between curve points manually placed along key plate edges and their orbital projections (**Figure 4**). Curves can be repositioned to simulate bending, trimming, or alternative placements. In the pilot study, five curves are placed (anterior floor, anterior medial wall, lateral floor, superior medial wall, floor–wall junction).

The module exports all raw distances and ranking summaries for patient-specific planning and data analysis (see the tutorial).

**C. Quantitative evaluation and visualization results**

Use of the patient data was approved from Baylor Research Institute Institutional Review Board (number 025-275). STL files of four preformed titanium orbital plates were obtained from two of the largest craniomaxillofacial hardware vendors (Vendor A and Vendor B), including small and large sizes from each vendor (**Figure 4**).

**C.1. Patient-Specific Fit Evaluation**

To illustrate patient-specific variability, four representative cases were examined to highlight the complex relationships between orbital morphology and plate geometry. Patient IDs were anonymized. Plate rankings are based on overall mean-edge specific distances.

**Patient 1048.** This patient presented with a major right orbital floor and minor orbital wall fractures. **The small Vendor A** plate achieved the best overall fit (mean distance = 0.998 mm). As shown in 2D and 3D views (**Figure 5**), its contour closely followed the reconstructed orbital floor and peripheral bone.

In contrast, both **Vendor B** plates exhibited a more convex posterior floor and more concave anterior floor, creating a high curvature at the floor component (**Figure 5**). This resulted in greater plate-to-orbit separation, particularly along the lateral posterior floor, which was also indicated by the blue-green regions in the heatmap (**Figure 5 and 6**). Notably, the anterior edge of the **small Vendor B plate** angled upward laterally, a unique feature that allowed it to better capture the anterior orbital rim (Rank 1 in anterior-floor edge fit; mean distance = 0.86 mm) (**Figure 5C and 5D**).

**Patient 1846.** This patient also had a right orbital floor fracture. The **small Vendor B plate** conformed well to the reconstructed orbit due to its convex posterior floor (rank 1, mean = 0.736 mm) (**Figure 7**). Its deep curvature and laterally upturned anterior edge also aligned well with the anterior orbital rim (**Figure 7C**). The flatter floor of the **large Vendor A plate** did not fully capture the posterior reconstructed floor contour when resting on the peripheral bone (rank 2, mean = 0.761 mm) (**Figure 7**).

**Patient 1633.** This patient presented with left orbital floor and medial wall fractures and had a relatively large orbit. The **large Vendor B** and **large Vendor A** plates ranked 1 and 2 (mean distances 0.98 and 1.20 mm, respectively). The more convex posterior floor and more concave anterior floor contours of the large Vendor B plate resulted in slightly better overall conformity (**Figure 8**). In contrast, the **small Vendor B plate** fits poorly with the anterior orbital rim of both the reconstructed and the original orbits (Rank 4, mean=1.884mm) (**Figure 8**).

**Cross-case summary.** Vendor A plates generally exhibit flatter contours, while Vendor B plates display a more curved posterior floor and distinctive anterior edge angulation. Notably, the small Vendor B plate, with its pronounced anterior curvature, showed good conformity in some smaller orbits. All these designs perform well in certain anatomical configurations and poorly in others. Collectively, these findings emphasize that optimal plate selection is highly patient-specific and cannot be determined by manufacturer and size alone.

## C.2. Summary of quantitative findings

Across the 14-patient pilot dataset, plate-wide conformity values ranged from approximately 1.4–1.7 mm across all plates, with similar average performance across vendors and sizes. Within each patient, however, one plate consistently achieved the lowest mean deviation, representing the best overall fit, even when differences between the top-ranked options were sometimes small. On average, the best-fitting plate showed markedly smaller deviations (≈1.1 mm) than the remaining plates (≈1.7 mm). These descriptive findings highlight substantial patient-specific variability in orbital morphology and reinforce the value of virtually comparing multiple plates during preoperative planning. This pilot study also demonstrates that the module generates standardized, exportable quantitative metrics suitable for downstream research. A larger follow-up study will enable more comprehensive statistical analysis.

## C.3. Collaborative planning, operator Biases, and educational application

The *PlateRegistration* module enables iterative adjustments of existing plate placements, facilitating collaborative review of alternative alignment strategies, plate modifications, plan sharing, and surgical education (see **Figure 9** for an example). This functionality also supports surgical education by allowing trainees and clinicians to explore how subtle changes in positioning affect plate-to-orbit conformity.

**Challenges**

**A. Data preparation and segmentation.**

As in other workflows based on 3D CT imaging, segmentation of fractured bone remains one of the most time-consuming steps but is essential for generating accurate 3D models for plate registration and fit evaluation. In this study, we used *DentalSegmentator*, a deep learning–based automated segmentation extension of 3D Slicer, to segment the skull and orbital bones. Manual segmentation using thresholding and editing can also be utilized. An experienced user can complete segmentation and orbital reconstruction in approximately 0.5–1 hour.

**B. Orbital reconstruction and effects of left-right asymmetry**

After segmentation, orbital reconstruction can be performed using the *MirrorOrbitalRecon* module. Mirrored-orbit reconstruction is widely used in both proprietary planning systems and academic studies, but natural left–right asymmetry introduces errors, and there is a lack of quantitative research on how this asymmetry affects reconstruction accuracy[1,2,5,8,16].

To address this, our workflow provides multiple reconstruction options. Beyond simple rigid alignment of the mirrored hemiskull or full skull, the *MirrorOrbitalRecon* module also provides an optional CPD deformable registration step to further reduce left–right deviations. Our qualitative evaluations suggest that minor asymmetry has little impact on plate-fit rankings, but future studies are needed to quantify these effects and evaluate whether such adjustments meaningfully improve the prediction of the original orbital contour.

**C. Fit metrics**

Plate-to-orbit distances represent a geometric approximation of conformity, but they rely on identifying the closest point on the orbit for each sampled point on the plate. Because this process is algorithmic and may not perfectly correlate with actual anatomy, we recommend only use mean distances across well-defined regions (e.g., edges of a plate) or across entire plates to measure conformity.

Plate-to-orbit distances are also influenced by differential operator decisions during plate placement, so operator-dependent differences can introduce bias. Future studies should evaluate the magnitude of operator-related variability and develop strategies to account for it in data analysis.

The current fit metrics are not comprehensive. A complete assessment of plate fit virtually will also require evaluation of plate bending and trimming, which is often done intraoperatively, and soft-tissue deformation in response to plate placement[1,2,6,8]. Such analyses require advanced physics-based simulations and will be essential for developing predictive models of surgical outcomes and hardware recommendations for orbital fracture repair[17–19]. Soft tissue and plate modification simulation is currently under development.

**D. Challenges in real surgical outcome evaluation**

Ultimately, assessing implant performance requires linking objective metrics with clinical outcomes, including plate stability, soft tissue position and volume, functional restoration, and risk of revision surgery[2,6,7]. In this pilot study, differences in plate-to-orbit conformity, although statistically significant in some comparisons, may not always translate into clinically meaningful differences. For example, in several cases the distinction between Rank-1 and Rank-4 plates was less than 1 mm, a magnitude that may not noticeably affect postoperative comfort or function.

Integrated geometric metrics with clinical outcomes is challenging for several reasons. Pre-injury and postoperative imaging data for the same patients are often unavailable, soft-tissue responses vary considerably across patients, and surgical techniques differ among operators and institutions. In addition, retrospective datasets rarely contain the detailed geometric information needed for quantitative analysis.

The virtual framework introduced in this work represents an initial but important step toward overcoming these limitations. It provides a reproducible, low-cost method for generating quantitative plate-conformity metrics as a foundation for investigating real clinical impact. Over time, this may enable more comprehensive evaluations of implant design, placement strategies, and patient-specific risk factors, goals that have previously been difficult to achieve using conventional virtual planning systems or retrospective imaging alone.

**Timeline**

The software is currently available to be installed on the most recent stable version of Slicer 5.10.0 via Extension Manager. It is readily usable with the provided sample data. In the upcoming year, we will begin to organize workshops to introduce this toolkit to surgical community, gather feedback, and initiate a community of developers and users to further developer this toolkit for both research and potential clinical innovation.

**Acknowledgement**

This study was funded by a Texas A&M Health Science Center Seedling Grant (awarded to CZ) and in part by the OMS Foundation. The authors thank the 3D Slicer community for their technical support and guidance throughout the software development process, particularly Kyle Sunderland (Perk Lab, Queen's University), Andras Lasso (Perk Lab, Queen's University), and Steve Pieper (Isomics Inc.).

**The use of AI and AI-assisted technologies statement**

Artificial intelligence–assisted technologies were used to improve the clarity and readability of the manuscript and to provide general programming assistance during software development (e.g., example code generation and debugging support). These tools were not used to generate scientific content, design study methodology, analyze or interpret data, or draw scientific conclusions. All scientific decisions, software architecture, coding, validation, and interpretation were performed by the authors, who take full responsibility for the accuracy, integrity, and originality of the work.

**Figures are presented at the end of the manuscript.**

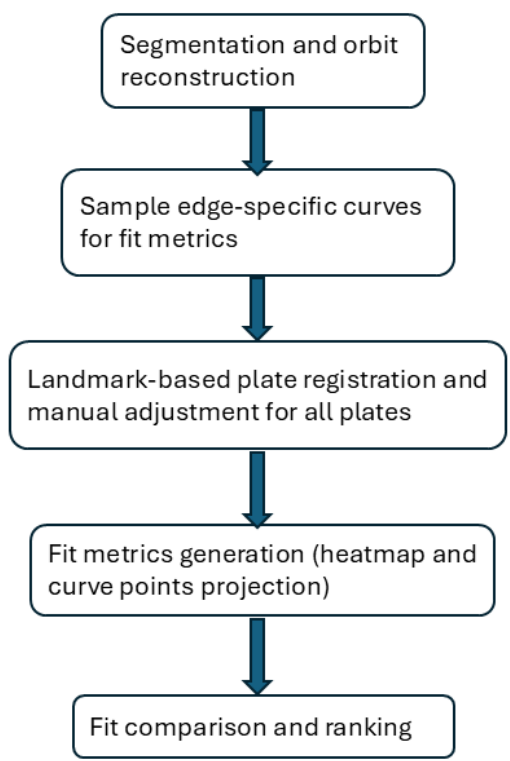

**Figure 1. Plate registration and fit comparison workflow**. See Appendix and tutorials for additional details.

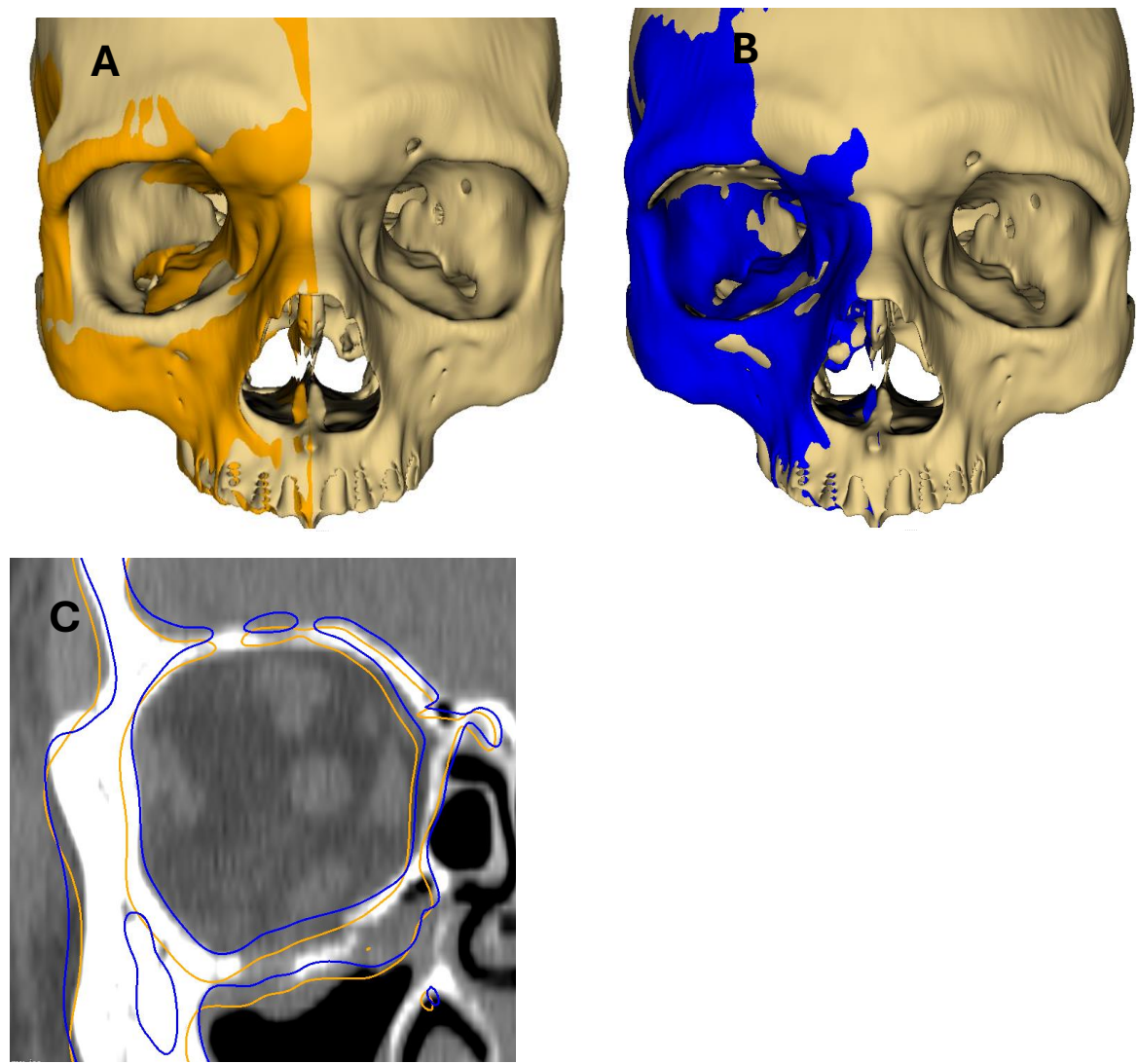

**Figure 2. Orbital reconstruction using the mirror of the contralateral hemiskull.** The skull is from patient ID 1846 (right orbital floor fracture). **(A)** Rigid registration of the reflected hemiskull (orange) in 3D view. **(B)** Affine registration of the reflected hemiskull in 3D view using a globa l affine transformation. **(C)** Superimposition of the rigid- and affine-registered hemiskulls in 2D slice view demonstrates that the globally affine-registered hemiskull (blue) achieves slightly better alignment with the intact orbital floor and medial wall than the rigidly registered one (orange). **Note that affine registration does not necessarily improve orbital alignment relative to rigid registration, as it applies a global transformation to the entire reflected hemiskull. The *MirrorOrbitRecon* module provides convenient options to select among different registration results to achieve the most accurate reconstruction.**

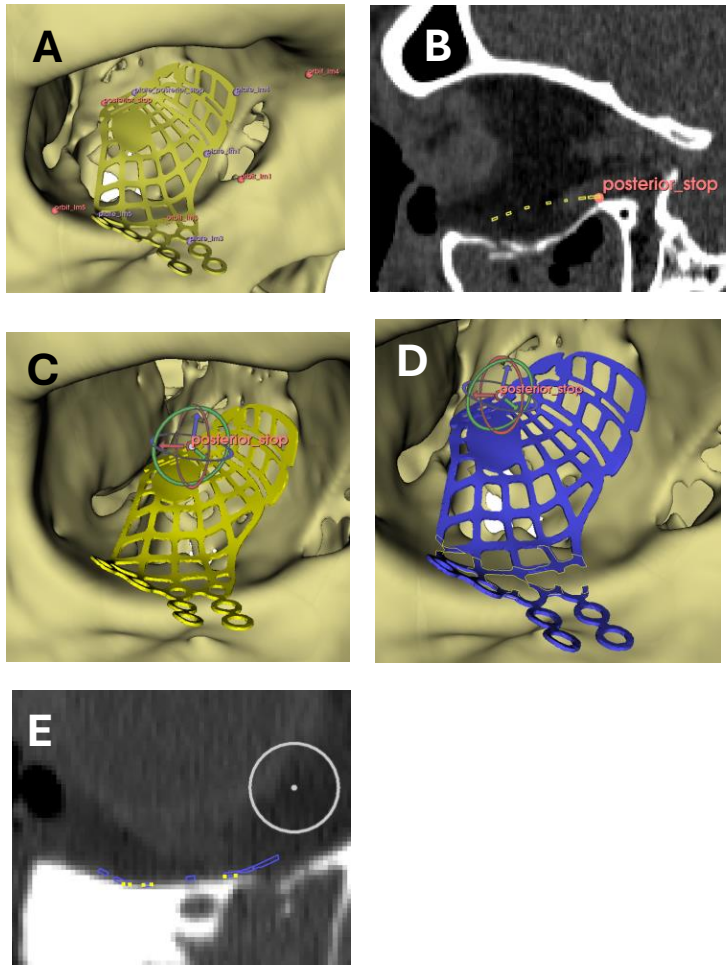

**Figure 3. Plate registration workflow. Figure S2. Plate registration workflow.**
**(A)** Initial landmark-based registration provides a rough alignment of the plate to the fracture site. **(B)** Posterior-stop alignment is then applied to anchor the plate at the posterior orbital stop. **(C)** Manual adjustment is performed using the Slicer's built-in 3D interaction handle (rotation is also enabled in 2D views) (developed by Kyle Sunderland, Perk Lab, Queens University). By default, the rotation center is fixed at the orbital stop, but users may optionally apply small translational adjustments. **(D)** Collision detection marks areas of contact in the 3D view (yellow lines). **(E)** Corresponding collision points are displayed in the 2D slice views (yellow dots). **(F)** The interface also reports the percentage of colliding plate area (see tutorial for details).

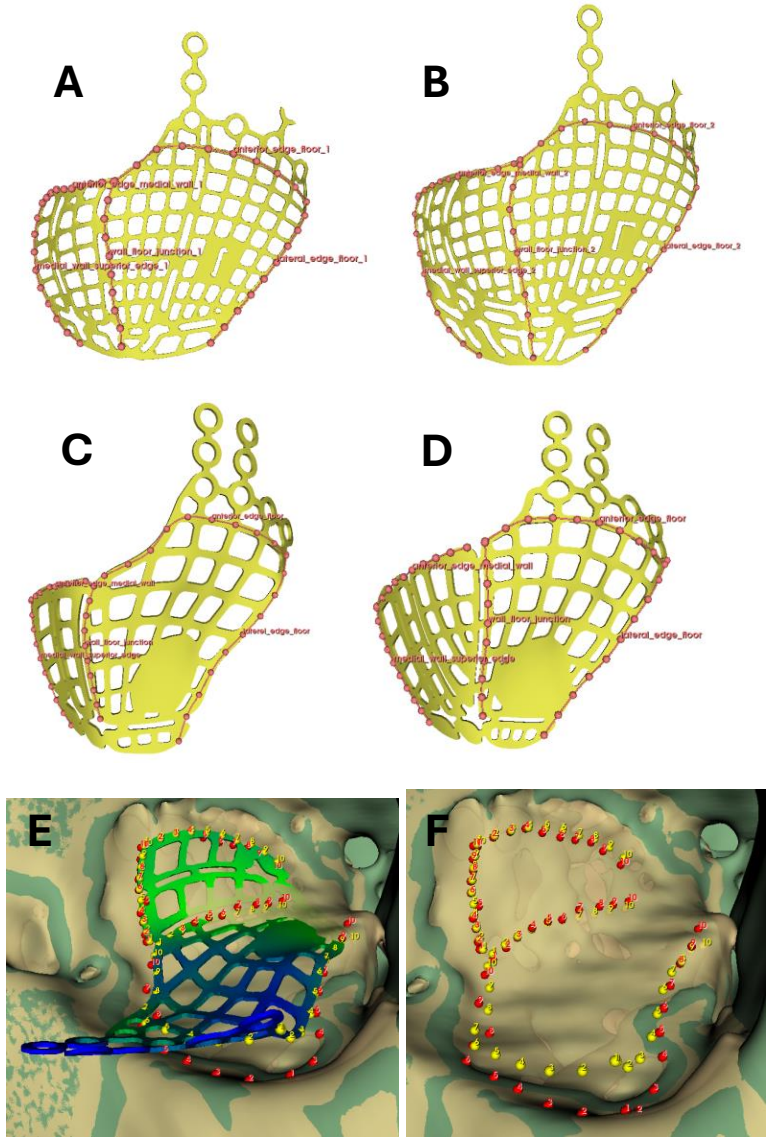

**Figure 4. Generate Edge-specific curve points and point projection.** For each plate, five curves were placed in Slicer to mark the five undersurface edges. Ten evenly spaced points were then resampled along each curve. **(A)** Left small Vendor A plate. **(B)** Left small Vendor B plate. **(C)** Large Vendor A plate. **(D)** Large Vendor B plate. **(E)** and **(F)** Point projection from the plate to the reconstructed orbit. Yellow points represent curve points on the plate edges, and red points represent the corresponding projected points on the reconstructed orbit (light yellow, semi-transparent surface). Distances are measured between curve points on the plate edges and their correspondent orbital projections.

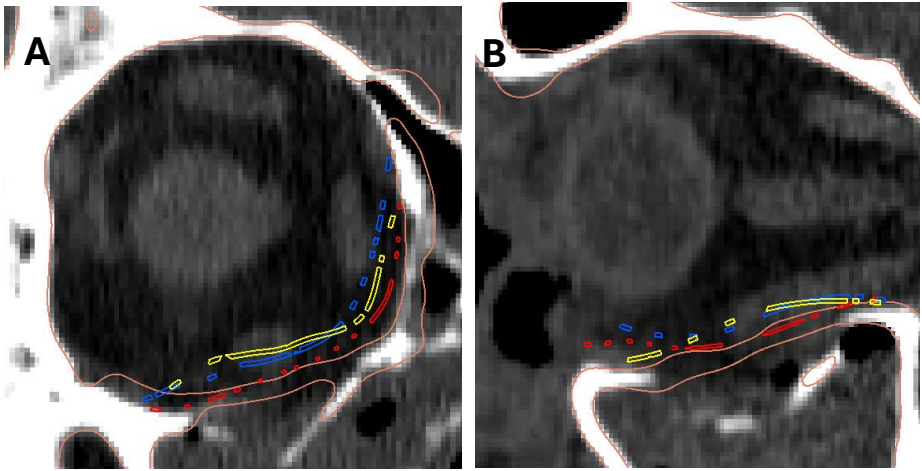
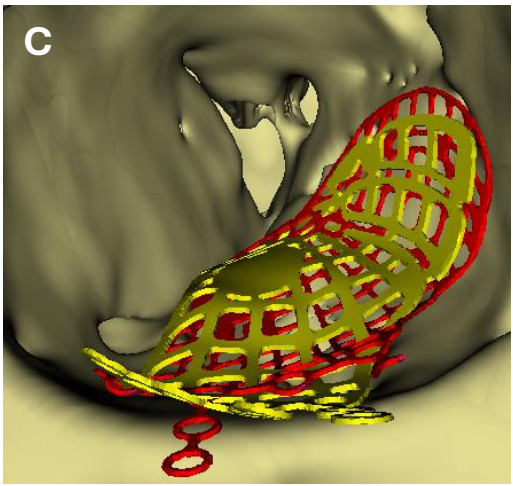
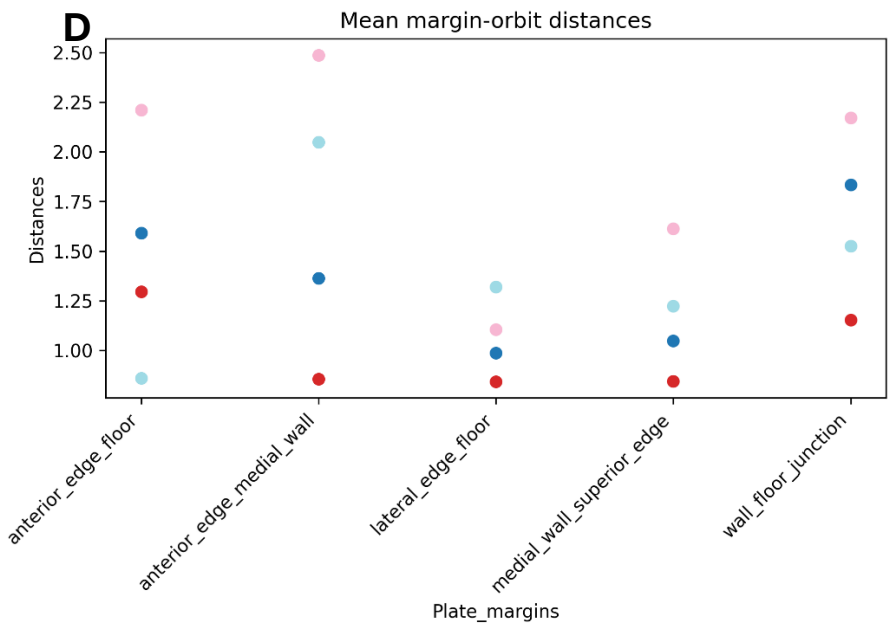

**Figure 5. Plate conformity assessment for Patient 1048.** Solid pink line: reconstructed orbit shown in CT slice views. Plate colors in 2D and 3D views: **Red = small Vendor A, yellow = small Vendor B, Blue = large Vendor B. (A)** Coronal and **(B)** sagittal slice views demonstrate that the flatter contour of the small Vendor A plate conforms more closely to the reconstructed orbital floor than either Vendor B plate. **(C)** In 3D view, the small Vendor B plate (yellow) show excessive posterior convexity relative to the small Vendor A plate (red). Notably, the anterior edge of the small Vendor B plate aligns well with the anterior orbital rim in this patient. **(D)** Scatterplot of mean edge-specific distances illustrates this finding: although Vendor B plates deviate more along most of the floor, the small Vendor B plate achieves the smallest distances at the anterior floor edge (**red dots: small Vendor A, light blue dots: small Vendor B, dark blue dots: large Vendor A, pink dots: large Vendor B**).

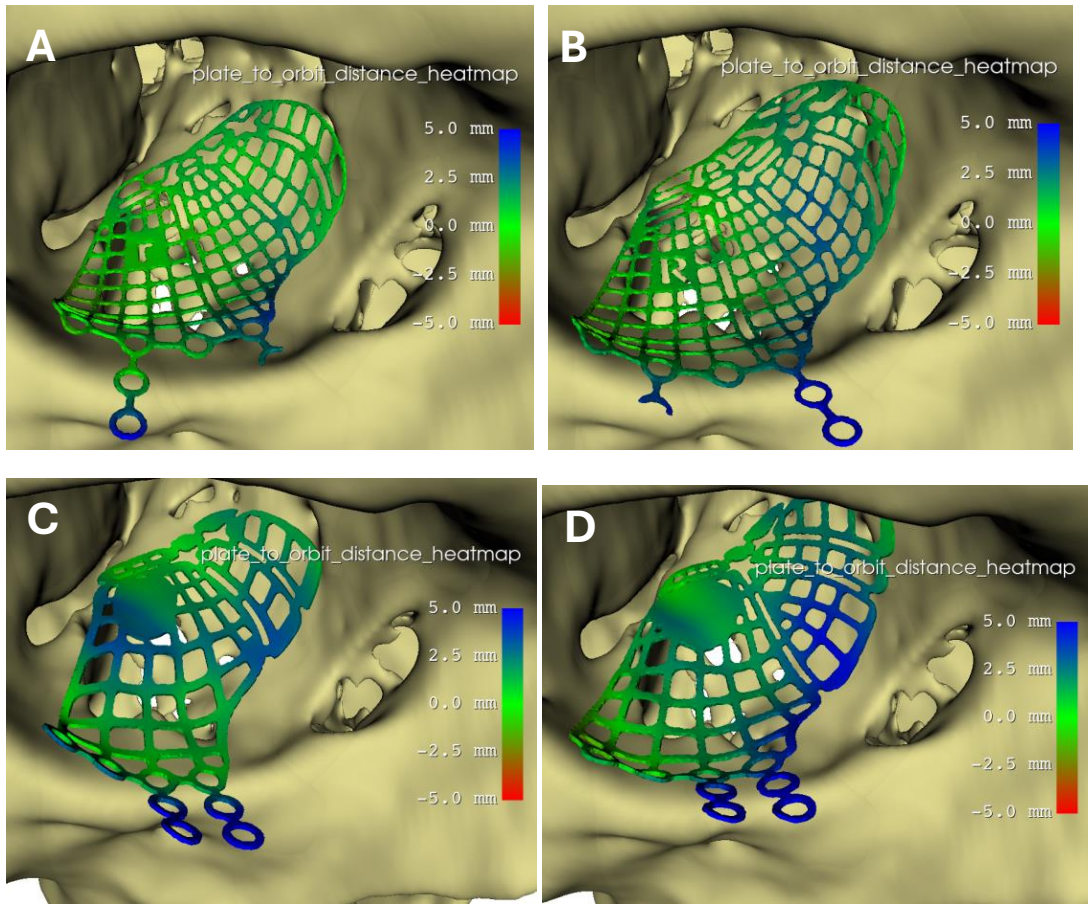

**Figure 6. Plate distance heatmaps (Patient 1048).** By default, heatmaps use a red–green–blue color scale representing plate-to-orbit distances from –5 mm (red) to +5 mm (blue) (reconstructed orbit was hidden for better visualization. Negative distances correspond to regions beneath the reconstructed orbit, whereas positive distances indicate regions above it. Users may adjust the scalar display range in the Slicer *Models* module to optimize visualization. Heatmap models, raw scalar values, and histograms are saved in the *fit_output/fit_metrics* folder within the user-specified directory for plate registration data. **(A)** Right small Vendor A plate. **(B)** Right large Vendor A plate. **(C)** Right small Vendor B plate. **(D)** Right large Vendor B plate. Heatmap generation is based on adapting functions from the Slicer *Model-to-Model Distances* module.

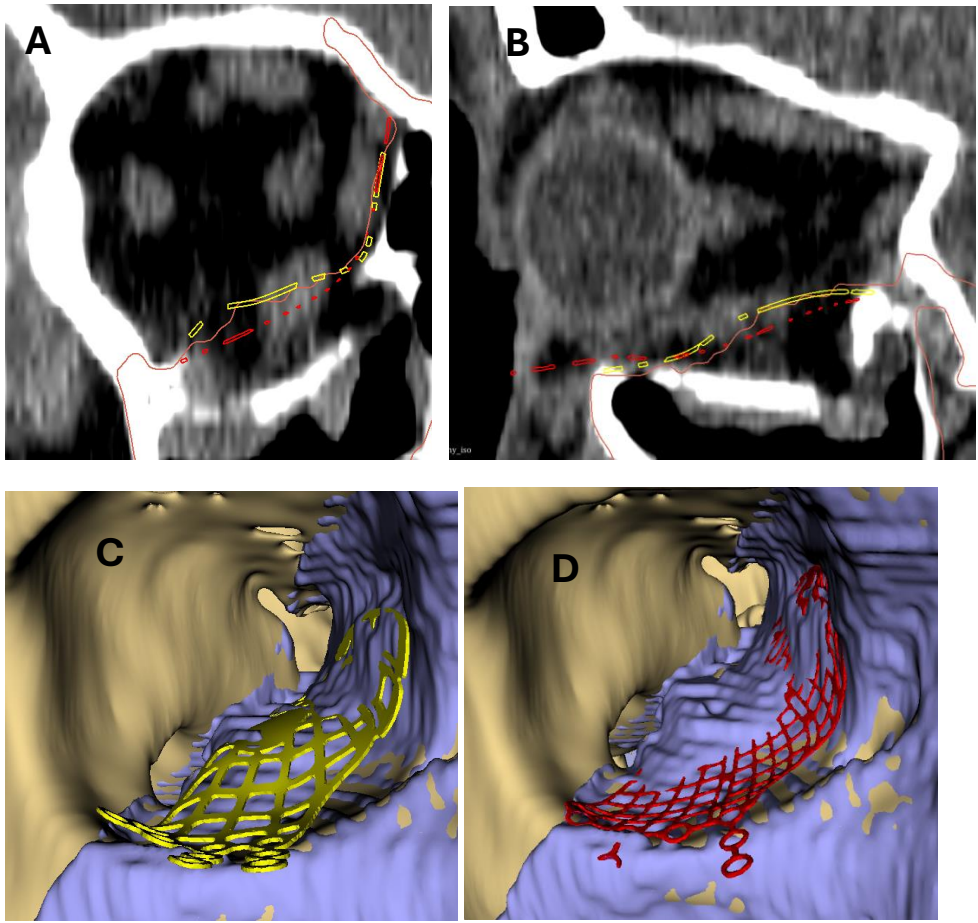

**Figure 7. Plate conformity assessment for Patient 1846.** Solid pink line: reconstructed orbit shown in CT slice views. Plate colors (2D and 3D views): **Red = large Vendor A, Yellow = small Vendor B. (A)** Coronal and **(B)** sagittal slice views demonstrate that the more convex floor contour of the small Vendor B plate (Rank 1, mean = 0.736mm) conforms more closely to the reconstructed orbital floor than the large Vendor A plate. **(C)** In the 3D view, the small Vendor B plate shows excessive posterior convexity relative to the small Vendor A plate. Notably, its anterior edge aligns well with the anterior orbital rim in this patient.

**(D)** Scatterplot of mean edge-specific distances illustrates this pattern: although Vendor B plates deviate more along most of the floor, the small Vendor B plate achieves the smallest distances at the anterior floor edge (red dots: small Vendor A; light blue dots: small Vendor B).

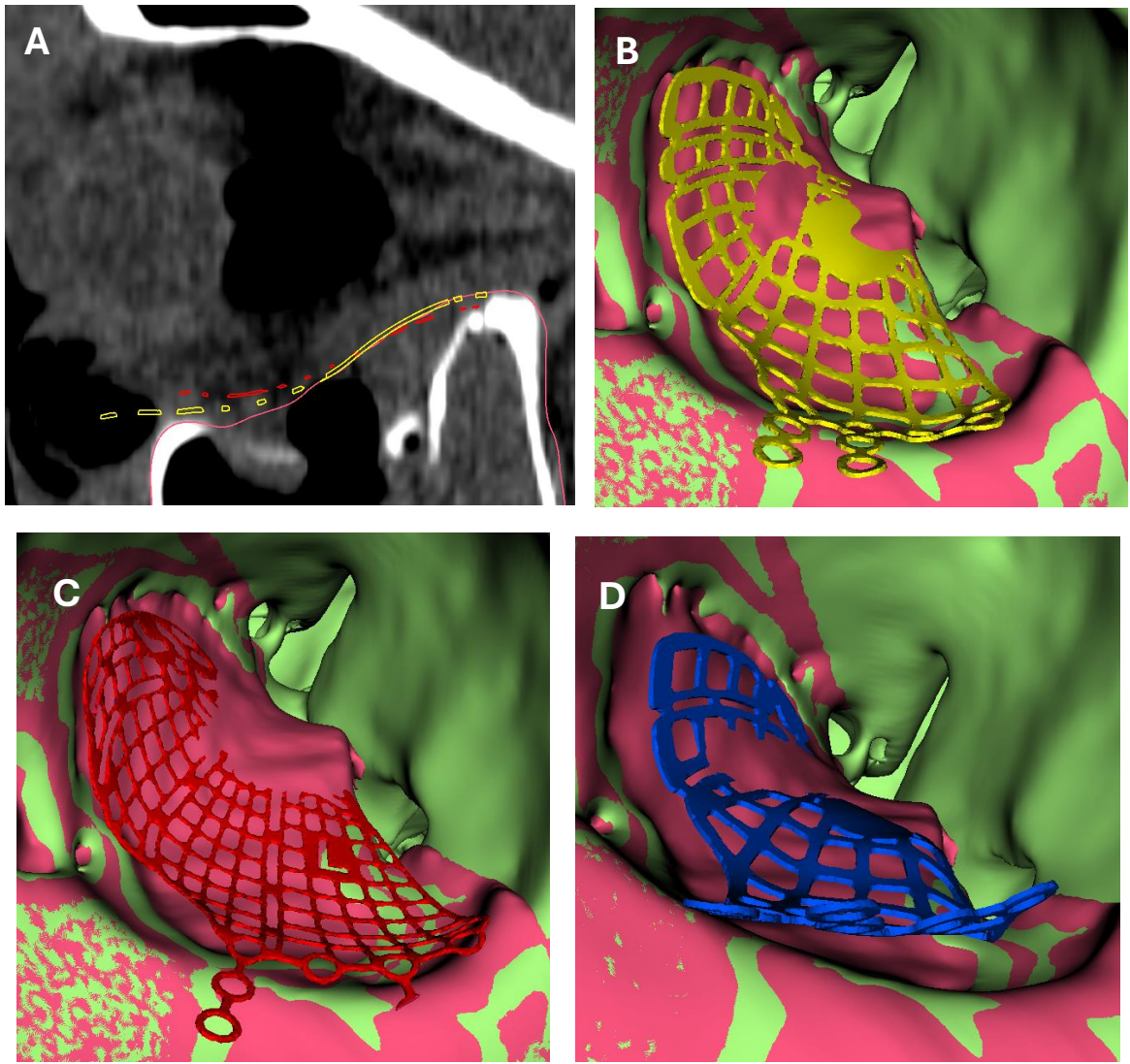

**Figure 8. Plate conformity assessment for Patient 1633.** Solid pink line = reconstructed orbit in CT slice views. Plate colors (2D and 3D views): **Red = large Vendor A, Yellow = large Vendor B, blue = small Vendor B. (A)** Sagittal slice view demonstrates that the more convex floor of the large Vendor B plate conforms better to the reconstructed orbit than the flatter large Vendor A plate.
**(B)** 3D view shows that the posterior floor of the large Vendor B plate aligns well with the reconstructed orbit (light red) (Rank 1, mean = 0.98 mm), while **(C)** the large Vendor A plate drops beneath the posterior reconstructed orbit, indicating poorer conformity (Rank 2, mean = 1.20 mm). **(D)** Although the small Vendor B plate also has a convex floor, its sharply angled anterior edge prevents proper adaptation to the orbital rim (Rank 4, mean = 1.88 mm).

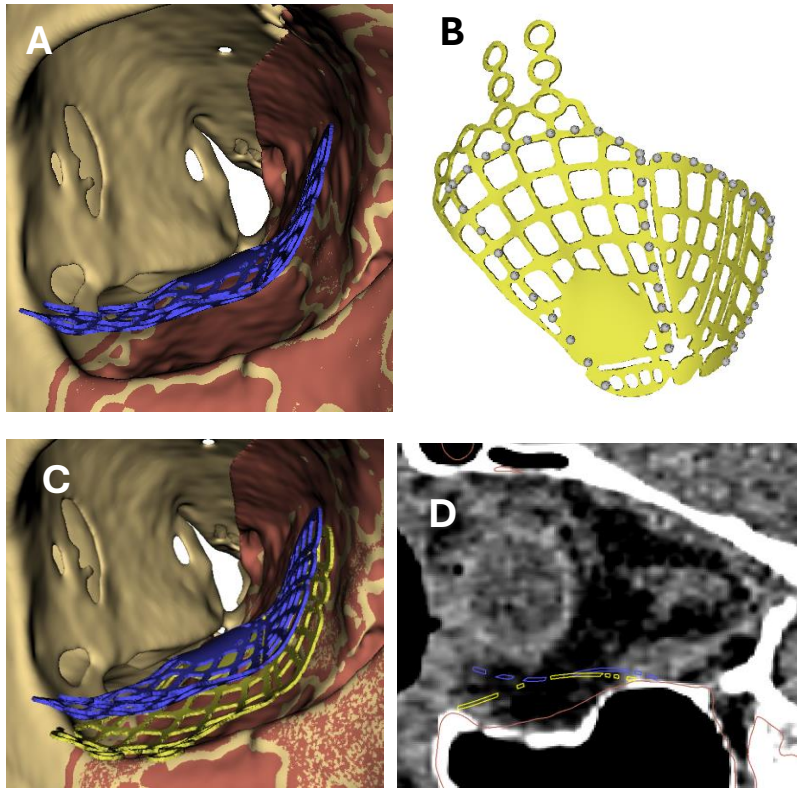

**Figure 9. Example of using SlicerOrbitSurgerySim to explore alternative plate-placement and trimming strategies (Patient 1937).**
**(A)** Original placement of the large Vendor B plate (blue) required superior lifting to avoid lateral-edge intersection, resulting in poor conformity (mean distance = 3.695 mm). **(B)** Simulated trimming was performed by repositioning the sampling curve medially to represent a shortened lateral margin, which allowed additional inferior and medial rotation during alignment. **(C)** 3D and **(D)** sagittal 3D views show that the modified plate position (yellow) substantially improved conformity, reducing the mean plate-to-orbit distance from 3.695 mm to 1.972 mm.

27